\documentclass[letterpaper, 10 pt, conference]{ieeeconf}  

\IEEEoverridecommandlockouts                              

\usepackage{amsmath,amsfonts,amssymb}
\usepackage{wrapfig}
\DeclareMathOperator*{\argmax}{argmax}
\usepackage{graphicx}
\usepackage{caption}
\usepackage{subcaption}
\usepackage{tensor}
\usepackage{algorithm}
\usepackage{algorithmic}
\usepackage{xcolor}
\usepackage{soul}
\usepackage{textcomp}

\newcommand{\norm}[1]{\left\lVert#1\right\rVert}

\title{\LARGE \bf
When Your Robot Breaks: Active Learning During Plant Failure
}

\author{Mariah L. Schrum$^{1}$ and Matthew C. Gombolay$^{1}$
\thanks{\copyright 2019 IEEE. Personal use of this material is permitted. Permission from IEEE must be
obtained for all other uses, in any current or future media, including
reprinting/republishing this material for advertising or promotional purposes, creating new
collective works, for resale or redistribution to servers or lists, or reuse of any copyrighted
component of this work in other works.}
\thanks{*This research is supported in part by the National Science Foundation (NSF) under Grant No. 1545287}
\thanks{$^{1}$Mariah L. Schrum and Matthew C. Gombolay are with The Institute for Robotics and Intelligent Machines,
        Georgia Institute of Technology, North Ave NW, Atlanta, GA 30332, USA
        {\tt\footnotesize \{mschrum3,mgombolay3\}@gatech.edu}}%
\thanks{}%
}

\begin{document}

\maketitle
\thispagestyle{empty}
\pagestyle{empty}

\begin{abstract}

Detecting and adapting to catastrophic failures in robotic systems requires a robot to learn its new dynamics quickly and safely to best accomplish its goals.  To address this challenging problem, we propose probabilistically-safe, online learning techniques to infer the altered dynamics of a robot at the moment a failure (e.g., physical damage) occurs. We combine model predictive control and active learning within a chance-constrained optimization framework to safely and efficiently learn the new plant model of the robot. We leverage a neural network for function approximation in learning the latent dynamics of the robot under failure conditions.   Our framework generalizes to various damage conditions while being computationally light-weight to advance real-time deployment. We empirically validate within a virtual environment that we can regain control of a severely damaged aircraft in seconds and require only 0.1 seconds to find safe, information-rich trajectories, outperforming state-of-the-art approaches.
\end{abstract}

\section{INTRODUCTION}

As robots increasingly become an integral part of our daily lives, our reliance on them grows accordingly. However, robots are susceptible to failure in the form of unexpected damage or routine wear and tear.  Even if a robot fails, the robot should be able to adapt to its new dynamics so that it can continue to function to mitigate the need for costly repairs or dangerous malfunctions.  For example, the motor of a bipedal robot may break mid-step, a tire may blow out on an autonomous car, or an actuator may fail on a UAV.  In such situations, there is a dual-need to try to maintain control given the robot's current understanding of its dynamics while also seeking out additional information to refine its model. These needs can be contradictory if seeking out information results in a terminal condition (e.g., a bipedal robot crashing to the ground). However, these needs can be complementary when ``safe" actions can be taken to gain new information about plant dynamics to better follow a desired trajectory.
	
The bounded rationality hypothesis describes how humans handle this cognitive dilemma of greedily operating under a known model of the world versus seeking out additional information specifically to refine this model \cite{Simon1990,Bendor2015}.  Bounded rationality refers to the theory that rationality in human decision making is limited by the tractability of the problem, available time, and cognitive resources.  When faced with these limitations, humans trade off between optimality of the solution and expenditure of resources, between information gain and executing actions efficiently. Robots, likewise limited by their computational resources, physical limitations under a failure condition, and time constraints, must make similar compromises.  Thus, when a robot experiences failure, it must use its resources to learn the nature of the failure efficiently to compensate in a timely fashion. Therefore, some knowledge of the full extent of the failure may need to be sacrificed to expedite reaching the goal. We model this failure-recovery problem as one in which the robot should focus on safely learning only the relevant aspects of its dynamics to efficiently accomplish the task at hand.

Prior work has sought to address safely learning a damage model~\cite{Bongard2006,Cully2015}. Bongard et al.~\cite{Bongard2006} and Cully et al.~\cite{Cully2015} demonstrate active learning methods to determine the true model of a damaged robot. However, these approaches are only effective when computational time is not a limiting factor. For example in \cite{Cully2015}, the robot took 66 seconds to learn how to operate after damage. This is far too slow in the case of an aircraft or other time constrained systems. To the best of our knowledge, no current architecture accounts for both a continuous distribution of damage and demonstrates the computational speed necessary to regain control of an unstable system, e.g., a damaged aircraft. While some approached have modeled failure dynamics from first principles, e.g.,~\cite{Baur2011,Hitachi2009}, these approaches are non-adaptive and restricted to a narrow set of point cases. For example, \cite{Hitachi2009 only considers the case of propulsion control for vertical tail damage and in \cite{Baur2011} the parameters of the controller are designed based on prior knowledge of the damage.} We move beyond the limitations of these prior works by developing active model learning techniques to safely acquire information about the altered plant dynamics to recover from failure.

In this paper, we contribute a novel chance-constrained, active learning, and model-based optimization algorithm to enable robots to efficiently and safely learn their new dynamics and recover from failure. Our bounded rationality framework trades off the risks of acquiring task-relevant information about the robot's failure dynamics with maximizing the probability that the robot can safely continue its mission. 
The active learning component of our algorithm mimics a human's need to seek additional knowledge when failure occurs. The safety framework counterbalances acquisition of new knowledge by optimizing for the goal under the current assumptions about the world. 
We contribute a novel acquisition function along with a powerful architecture for learning and controlling a damaged robot in real time. We empirically validate within a virtual environment that we can regain control of a severely damaged aircraft in seconds and require only 0.1 seconds to find safe, information-rich trajectories, outperforming state-of-the-art methods.

\section{Related Works}
We draw upon research in Model Predictive Control (MPC), Active Learning, and recovery from failure to create a novel architecture allowing a robot to safely recover in real time from a large distribution of damage scenarios.

MPC utilizes a model of the plant to make predictions about the plant's future behaviors and approximate the optimal control based on these predictions \cite{Mayne2002,CarlosE.1989}.  Recent work has shown the potential for MPC to be used in conjunction with online learning techniques.  \cite{Wagener2019} presents an online learning approach to designing model predictive controllers.  This framework utilizes online learning techniques to learn the parameters that minimize the MPC objective. The authors demonstrate their algorithm's capabilities on a driving task. \cite{Williams2017} proposes using neural networks as dynamic models in an MPC scheme. A sampling-based, information theoretic algorithm is proposed to optimize the MPC cost function.

Active Learning attempts to address the problem that training data is often expensive to obtain and label.  Knowledge about which training inputs provide the most information to the algorithm, if their labels are known, is often highly useful. Active Learning has been studied in the context of supervised learning and classification \cite{Hasenjager1998,Yang2015} and  regression \cite{Burbidge2007, Diagrams2010}. Several previous approaches have employed active learning for model learning. \cite{Bongard2006} demonstrates an active learning method to learn a damage model by generating candidate models and using active learning to select the most likely model. \cite{Cully2015} utilizes active learning and a Gaussian process model to learn a damage model.

Detection of and recovery from failure has been studied extensively in aircraft  \cite{Bakshi2018}. \cite{Baur2011} proposed linear equations of motion for an aircraft suffering from wing damage and actuator damage and implements a model reference adaptive controller to compensate for these failures.  The researchers demonstrated they could accurately track a reference.
\cite{Hitachi2009} proposed a propulsion-only controller using H-infinity loop transfer recovery to control a plane that has suffered loss of hydraulic function.  The authors demonstrated the validity of using an H-infinity controller in several damage cases. While effective in certain situations, these approaches often rely on prior knowledge of the damage and do not generalize well to a large space of possible damage conditions.

\section{Motivating Application: Aviation}
We motivate the need for robots to operate under failure in the problem of aircraft recovery from damage. Aircraft are susceptible to a range of failure scenarios, which are difficult to predict and model, and have tight time constraints for collecting data. 
For example, 260 lives were lost when the rudder of American Airlines Flight 587 snapped off and the pilot could not recover control \cite{NTSB}. In 2005, the wing of Chalk's Ocean Airways Flight 101 broke off due to structural weakness resulting in the death of all passengers \cite{NTSBa}. 

\begin{figure}
    \centering
    \includegraphics[width = 0.6\linewidth]{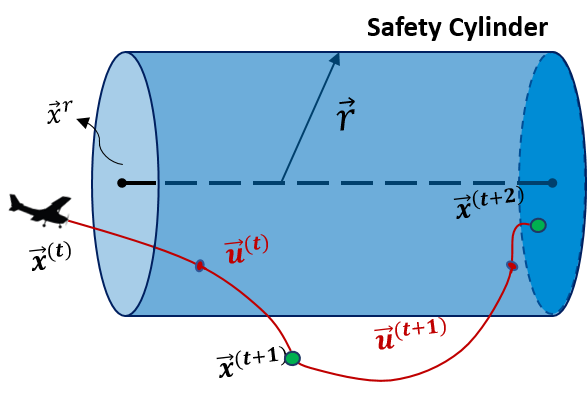}
    \caption{This figure depicts our objective to take an action to acquire information while having a high probability of returning to the safety envelope.} 
    \label{fig:safetyDiagram}
\end{figure}

Our objective is two-fold: 1) given a safe aircraft configuration envelope, take actions that have a high probability (e.g.,  $p > 1-\epsilon$) that the aircraft is able to return to a safe flight envelope and 2) maximize information gain along the aircraft's trajectory out of the envelope. \par\nobreak{\parskip0pt \footnotesize \noindent\begin{equation}
\vec{\mathcal{U}}^{(t:T)^*} =\argmax_{\vec{\mathcal{U}}^{(t:T)} \in \boldsymbol{\vec{\mathcal{U}}}^{(t:T)}}  I(\vec{\mathcal{U}}^{(t:T)})+\lambda \text{Pr}\Big\{ \norm{\vec{x}_{t+T} - \vec{x}^r}_1 \leq r \Big\} \label{eq:7}
\end{equation}}\indent The trade off between safe flight and information gain is described in Eq.~\ref{eq:7}.  Here, $I(\vec{u})$ is a measure of the amount of information gained when taking action $\vec{u}$. $\lambda$ is a parameter that can be adjusted depending on the desired trade off between learning and the probability of remaining in a safe region.  $\vec{x}^r$ is the safe reference trajectory and r is the radius of the cylinder of safety in which the robot can explore. This safety cylinder defines the configurations that are safe for the robot to be in. The probability that the robot can return to the cylinder of safety after taking action $u^{(t)}$ via action $\vec{u}^{(t+T)}$ is to be maximized in light of the desire to also maximize information gained about the robot's dynamics along the trajectory from $[t,t+T)$. $\mathcal{U}$ is the set of possible actions. This formulation, as defined in Eq.~\ref{eq:7}, ensures that maximal information is gained in each time step while seeking a high probability that the aircraft will return to the safe cylinder. For convenience, we define $\vec{\mathcal{U}}^{(t:T)} = \begin{bmatrix} \left[\vec{u}^{(t)}\right]^\intercal,\ldots, \left[\vec{u}^{(t+T)}\right]^\intercal \end{bmatrix}^\intercal$ and $\boldsymbol{\vec{\mathcal{U}}}^{(t:T)}$ as the set of such action trajectories. A visualization of our objective is shown in Fig.~\ref{fig:safetyDiagram} for the case of $T=2$.

\section{Algorithmic Overview}
\label{sec:overview}
We present our closed-loop learning mechanism below. Before damage occurs, the robot follows a MPC policy, assuming a nominal plant model.  At each time step $t$, given action $u^{(t)}$, we monitor the predicted plant output $\vec{x}^{(t+1)}$ provided by the nominal plant model and compare it to the actual measurement of the system. If the error between the predicted measurement and the true measurement is above a threshold, we assume a mismatch between our nominal plant model and the true dynamics of the system, meaning damage may have occurred. The goal is now to learn a model that represents the change between the nominal dynamics and the damage dynamics. The decision to learn the change in the dynamics was inspired by the previous work of \cite{Kory,Saver} which showed that learning model displacements is more effective than re-learning a model from scratch.
 
Once the mismatch has been detected, the robot explores to learn more about the nature of the damage by collecting a set of training examples consisting of $u^{(t)}$, $x^{(t)}$, and $x^{(t+1)}$ (Line 2).  We train a single-layer perceptron and utilize the acquisition functions discussed below to determine the next action to take that maximizes our active learning metric (Lines 4-5).  The action must also satisfy the condition that it is ``safe".  This means that with probability $1-\epsilon$, the robot will be able to return to a safe state. We continue taking safe actions determined via optimization of the active learning acquisition function to improve upon the preliminary damage model as quickly as possible.  Once our confidence in the model reaches a satisfactory level, we update the plant model utilized by our MPC  (Line 7-8). We continue refining the model over time as new data is acquired. Confidence is a function of the active learning metric.
\begin{algorithm}
\begin{algorithmic}[1]
\WHILE{true}
\IF{error detected}
\STATE $\vec{u}_{i...N},\vec{x}_{i...N} \leftarrow sensors$
\WHILE{error above threshold}
\STATE \footnotesize $u^*=\argmax_{u \in \mathcal{U}} I+\lambda \text{Pr}\Big\{ \norm{\vec{x}_{t+T} - \vec{x}^r}_1 \leq r \Big\}$
\STATE update $f(x_{i...N},u_{i...N})$
\ENDWHILE
\ELSE
\STATE MPC Plant Model $\leftarrow f$
\STATE $u^* \leftarrow$ MPC policy
\ENDIF
\ENDWHILE
\end{algorithmic}
\caption{Overview of our safe learning framework.}
\label{alg:Overview}
\end{algorithm}


In summary, if damage has been detected, the robot follows a policy provided by the bounded rationality framework, safely exploring the environment to efficiently learn the updated model. Once the model has been learned, the robot follows a nominal MPC policy, utilizing the updated plant model. In the next section, we present our novel optimization approach to probabilistically-safe active learning to adapt to failure in robotic systems (Sec.~\ref{sec:ourDesc}). Finally, we incorporate neural network function approximation (Sec.~\ref{sec:FA}) within our mathematical optimization and lightweight, high-quality active learning acquisition functions (Sec.~\ref{sec:AL}). 

\section{Bounded Rationality Framework}
\label{sec:ourDesc}
We formulate the problem of bounded-rationality control of robots during failure as a probabilistic, mixed integer linear program, as shown in Eq.~\ref{eq:obj}-\ref{eq:safety}, which is solved using a commercial solver employing a branch and bound method. We adopt the same definition from Eq.~\ref{eq:7} in which the robot trades off the information it could gain to improve the system's controllability while also trying to achieve a high-probability of safe-flight by staying within a specified safety envelope. Our objective function is defined in Eq.~\ref{eq:obj} in which we optimize a finite-horizon trajectory over $t' \in \{t,t+1,\ldots,t+T\}$ to maximize a trade off between our information gain, $I(\vec{u})$, and our safety goal, $g(\vec{u})$. 

Information gain, $I(\vec{u}^{k})$, is formulated (Eq.~\ref{eq:I}) as the inverse of the similarity between candidate data and previous training examples, where $N$ is the number of stored data points included in our analysis. Our novel acquisition function is presented in comparison to state-of-the-art functions in \ref{sec:AL}. Our safety objective, $g(\vec{u})$, is defined in Eq.~\ref{eq:safety}. The probability of safety is a conjunction of each dimensions, $d$, of safety envelope, $\vec{r}$, with a time-varying center at $x_{t+T,d}^r = h(d,t+T)$. Our dynamics are given by $\vec{x}^{(k+1)} = f(\vec{x}^{(k)},\vec{u}^{(k)})$ and are not necessarily linear.  \par\nobreak{\parskip0pt \footnotesize \noindent
\begin{gather}
    \vec{\mathcal{U}}^{(t:T)^*}= \argmax_{\vec{\mathcal{U}}^{(t:T)} \in \boldsymbol{\vec{\mathcal{U}}}^{(t:T)}} \sum_{k = t}^{t+T} I(\vec{u}^{(k)})  + \lambda g(\vec{u}^{(t+T)})\label{eq:obj} \\
    I(\vec{u}^{k}) = \sum_{i=1}^N \norm{u^{(k)}_d-u^{(i)}_{d}}_1+\beta\norm{f\left(\vec{x}^{(k)},\vec{u}^{(k)}\right)-x^{(i)}_{d}}_1 \label{eq:I}\\
    g(\vec{u}^{(t+T)}) =  \text{Pr}\Big\{\land_{d=1}^D \Big(\norm{x^{(t+T)}_d-x_{d}^r}_1<r_d\Big)\Big\}
    \label{eq:safety}
\end{gather}}This mathematical program is a linear-objective, nonlinearly-constrained optimization problem. In particular, the absolute values in $I(u)$ from Eq.~\ref{eq:I} and inside the probability in Eq.~\ref{eq:safety}, both impart piece-wise linearities and the $d-$conjunction of envelope-satisfaction events introduces a $d-$degree polynomial form. Unfortunately, modern solvers are not readily able to handle the non-convexities introduced by these constraints.

To gain computational tractability, we derive a novel linearization that affords sub-second optimization of the trajectory as shown in Eq.~\ref{eq:acqLin1}-\ref{eq:acqLin3}. This formulation is able to accomplish sub-second optimization while maintaining information-rich, probabilistically-safe trajectories, which we empirically demonstrate in Sec.~\ref{sec:ultimateEval}. Our first step is to transform each piece-wise term in our acquisition function into a set of integer, linear constraints, as shown in Eq.~\ref{eq:acqLin1}-\ref{eq:acqLin5}, where $M$ is a large positive number, $z_{d}^{(k,i)},\zeta_{d}^{(k,i)} \in [0,\infty)$, and $\pi^{(k,i)}_d,\nu^{(k,i)}_d  \in \{0,1\}$. This ``big M" method~\cite{martin1987generating} in Eq.~\ref{eq:acqLin2}-\ref{eq:acqLin3} makes one of the two inequalities mute when the integer variable in the corresponding equation takes on the value of zero. While this introduces $O(N^2D)$ integer variables, we show in Sec.~\ref{sec:ultimateEval} that we solve this problem in $< 1$ second.\par\nobreak{\parskip0pt \footnotesize \noindent
\begin{align}
I(\vec{u}^{(k)}) &= \sum_{i=1}^N \sum_{d=1}^D z_{d}^{(k,i)} +\beta \zeta_{d}^{(k,i)}, \forall k \label{eq:acqLin1}\\
\zeta_{d}^{(k,i)}&\leq x^{(k+1)}_d-x_{d}^{(i)}+M\left(1-\pi^{(k,i)}_d\right),\forall i,j,k \label{eq:acqLin2}\\
\zeta_{d}^{(k,i)}&\leq x_{d}^{(i)}-x^{(k+1)}_d+M\pi^{(k,i)}_d,\forall i,j,k \label{eq:acqLin3}\\
z_{d}^{(k,i)}&\leq u_{d}^{(k)}-u_{d}^{(i)}+M\left(1-\nu_d^{(k,i)}\right), \forall i,j,k \label{eq:acqLin4} \\
z_{d}^{(k,i)}&\leq  u_{d}^{(i)}-u_{d}^{(k)}+M\nu_d^{(k,i)}, \forall i,j,k \label{eq:acqLin5}
\end{align}}\indent The next step is to linearize Eq.~\ref{eq:safety}. First, assume the dynamics are piecewise-linear (e.g., as one would find in a neural network function approximator with a mixture of rectified linear units (ReLU) and linear activation functions). Second, we assume our model error comes from a Gaussian distribution with a known mean and variance. We leave for future work reasoning about the model's meta uncertainty (i.e., error in the estimates for the mean and variance).

For simplicity but without loss of generality, we consider a derivation of our dynamics for a two-step horizon (i.e., $T = 2$) and a neural network with linear activations for approximating the plant dynamics. Under these conditions, we would have Eq.~\ref{eq:midway1} from which we wish to enforce constraint Eq.~\ref{eq:midway2}. $\land_{d=1}^D$ is the logical conjunction of associated predicates indexed by $d$. In our context, $\text{Pr}\Big\{\land_{d=1}^D \Big\}$ is the probability of all events indexed by d occurring as true. In the following sections $\norm{x}$ refers to the L-1 norm and $|x|$ refers to the absolute value of $x$.\par\nobreak{\parskip0pt \footnotesize \noindent
\begin{gather}
    \vec{x}^{(t+2)}=(A^2+2A+I)\vec{x}^{(t)}+(AB+B)\vec{u}^{(t)} + B\vec{u}^{(t+1)} \label{eq:midway1}\\
    g(\vec{u}^{(t+2)}) = \text{Pr}\Big\{\land_{d=1}^D \Big(\norm{x_{d}^{(t+2)}-x_{d}^r}_1<r_d\Big)\Big\}  \label{eq:midway2}
\end{gather}}\indent Under a Gaussian assumption of the dynamics, as captured by matrices $A$ and $B$ in Eq.~\ref{eq:midway1}, we can then re-write the equations directly capturing this probability, as shown in Eq.~\ref{eq:midway3a} with $g(\vec{u}^{(t+2)}) = 1-\epsilon_d$. Here, $\bar{A}$ and $\bar{B}$ are the point estimates of the dynamics as predicted by the function approximator and $\bar{a_d}$ and $\bar{b_d}$ are rows of the associated matrices. $\sigma$ is the matrix of the associated standard deviations of the weights. $\Phi$ is the cumulative distribution function (CDF) for the normal distribution. Indices $d$ and $j$ indicate rows and columns of the associated matrices, $1-\epsilon_d$ is the probability level, and $\Delta_d^{(t:T)} =x^r_d-(\bar{a}^2_d+2\bar{a_d}+1)x^{(t)}_d$.\par\nobreak{\parskip0pt \footnotesize \noindent
\begin{align}
    &\Phi^{-1}(1-\epsilon_d)\sqrt{\sum_{j}\sigma_{d,j}^2{x}_j^{(t)^2}+\sum_{j}\sigma^{^2}_{d,j}\mathcal{U}^{(t:T)^2}_{j}}\nonumber \\ 
    &\indent\indent\indent\indent+ \begin{bmatrix}\bar{ab_d}+\bar{b_d} & \bar{b_d}
         \end{bmatrix} \vec{\mathcal{U}}^{(t:T)^2} <  r_d+\Delta_d^{(t:2)}\label{eq:midway3a} \\
         &-\Phi^{-1}(1-\epsilon_d)\sqrt{\sum_{j}\sigma_{d,j}^2{x}_j^{(t)^2}+\sum_{j}\sigma^{^2}_{d,j}\mathcal{U}^{(t:T)^2}_{j}}\nonumber \\ 
    &\indent\indent\indent\indent- \begin{bmatrix}\bar{ab_d}+\bar{b_d} & \bar{b_d}
         \end{bmatrix} \vec{\mathcal{U}}^{(t:T)^2} <  r_d-\Delta_d^{(t:2)}\label{eq:midway3b}
\end{align}}\indent The challenge lies in that the square root of the sum of squares is nonlinear and that the CDF of the normal distribution lacks an analytical inverse. For the sum of squares, we make the conservative assumption that {\footnotesize $0\leq \sqrt{\sum_{j} \sigma_j^2x_j^2} \leq \sum_{j}\sigma_j|x_j|$}. For the CDF, we adopt a ``probability-selector variable,'' $\delta_{\epsilon_{p,d}}$, which is $0$ when describing the probability of satisfying the constraint for dimension $d$ with probability $p$ and $1$ when the constraint is ignored, which allows us to replace the CDF call with a constant value for $p$. These augmentations yield $g(\vec{u}_{t+T}) = (1-\delta_{p,d})(1-\epsilon_p)$ described by Eq.~\ref{eq:safetyLin1}-\ref{eq:safetyLin3}, where $E$ is the set of ``probability levels'' allowed, e.g., $E = \{0.05,0.04,...\}$, and $\epsilon_{p,d} \in \{0,1\}, \forall p \in E, d \in D$. We also bound $\vec{\mathcal{U}}^{(t:2)}$ based on the range of possible inputs that can be achieved by the system.\par\nobreak{\parskip0pt \footnotesize \noindent
\begin{align}
&-M\delta_{\epsilon}-\Phi^{-1}(1-\epsilon_p)\sum_{j}\sigma_{d,j}\mathcal{\tilde{U}}_j^{(t:2)} \nonumber \\
&\indent - \begin{bmatrix}\bar{ab_d}+\bar{b_d} & \bar{b_d}
         \end{bmatrix}\vec{\mathcal{U}}^{(t:2)}<\vec{r_d}+\Delta_d^{(t:2)}+\sum_{j}\sigma_{d,j}|{x^{(t)}_j}| \label{eq:safetyLin1} \\
&-M\delta_{\epsilon}+\Phi^{-1}(1-\epsilon_p)\sum_{j}\sigma_{d,j}\mathcal{\tilde{U}}_j^{(t:2)} \nonumber \\
&\indent + \begin{bmatrix}\bar{ab_d}+\bar{b_d} & \bar{b_d}
         \end{bmatrix} \vec{\mathcal{U}}^{(t:2)}<r_d-\Delta_d^{(t:2)}-\sum_{j}\sigma_{d,j}|{x_j^{(t)}}| \label{eq:safetyLin2} \\
&\sum_{p \in E}\delta_{\epsilon_{p,d}}=|E|-1, \forall d \in D \label{eq:safetyLin3}
\end{align}}

\subsection{Neural Network Function Approximation}
\label{sec:FA}
Our derivation until now (i.e., Eq.~\ref{eq:midway1}, \ref{eq:midway3a}-\ref{eq:safetyLin2}) has assumed a linear, continuous dynamics model in the form of $\vec{x}_{t+1} = A\vec{x}_t + B\vec{u}_{t}$. Specifically, we have assumed that a multiple linear regression model is used to learn $A$ and $B$ each time the optimization problem is solved. However, the dynamics of an aircraft  are often highly nonlinear, thus requiring a more sophisticated function approximator. We draw support from the universal function approximation theorem for width-bounded networks with ReLU activations~\cite{lu2017expressive}.

We re-derive our model for a two-layer neural network, with ReLU activations in the first layer and a fully-connected layer for the second, as shown in Eq.~\ref{eq:ReLU-1}-\ref{eq:ReLU-3}, where $\tensor[^{(l)}]{o}{_i}$ is the output of neuron $i$ in layer $l$, $\tensor[^{(l)}]{\omega}{_{i,j}}$ is the connection from neuron $i$ in layer $l$ to neuron $j$ in layer $l+1$, and $\Xi^{(t)} = \left[
\left[x^{(t)}\right]^\intercal,\left[u^{(t)}\right]^\intercal\right]^\intercal$. 

{\small
\begin{align}
\hat{x}^{(t+1)}_d &= \sum_i  \tensor[^{(2)}]{\omega}{_{j,d}} * \tensor[^{(2)}]{o}{_i}, \forall d \in D \label{eq:ReLU-1}\\
\tensor[^{(2)}]{o}{_i} &= \sum_j  \tensor[^{(1)}]{\omega}{_{j,i}} * \tensor[^{(1)}]{o}{_j} \label{eq:ReLU-2}, \forall i\\
\tensor[^{(1)}]{o}{_i} &= \left\{ \begin{array}{cc} 
             \sum_j  \tensor[^{(0)}]{\omega}{_{j,i}} \Xi^{(t)}_j & \text{ if }  \sum_j \tensor[^{(0)}]{\omega}{_{j,i}} \Xi^{(t)}_{j} > 0 \\
            0 & \text{otherwise}  \\
            \end{array} \right. \label{eq:ReLU-3}
\end{align}}

We must then transform this final equation into mixed-integer linear constraints to fit within our optimization framework, as shown in Eq.~\ref{eq:ReLU-Lin1}-\ref{eq:ReLU-Lin6}, with $\tensor[^{(1)}]{o}{_i}\geq 0, \forall i$. \par\nobreak{\parskip0pt \footnotesize \noindent
\begin{align}
 M\xi_i -M + \sum_j\tensor[^{(0)}]{\omega}{_{j,i}} \Xi^{(t)}_j \leq 0 &\leq M \xi_i + \sum_j\tensor[^{(0)}]{\omega}{_{j,i}} \Xi^{(t)}_j \label{eq:ReLU-Lin1}\\
\sum_j\tensor[^{(0)}]{\omega}{_{j,i}}\Xi^{(t)}_j - M \leq \tensor[^{(1)}]{o}{_i}&\leq \sum_j\tensor[^{(0)}]{\omega}{_{j,i}}\Xi^{(t)}_j + M \xi_i \\
M - M \xi_i \geq \tensor[^{(1)}]{o}{_i}&\geq  \xi_i , \forall i \label{eq:ReLU-Lin6}
\end{align}}Finally, we can incorporate our mixed-integer linear formulation of a ReLU neural network within our dynamical equations (i.e., Eq.~\ref{eq:safetyLin1}-\ref{eq:safetyLin2}). Instead of propagating with the linear formulation from Eq.~\ref{eq:midway1}, we employ a recursive set of Eq.~\ref{eq:ReLU-1}-\ref{eq:ReLU-3} for each time step.

\begin{figure}
    \centering
    \includegraphics[width=0.95\columnwidth]{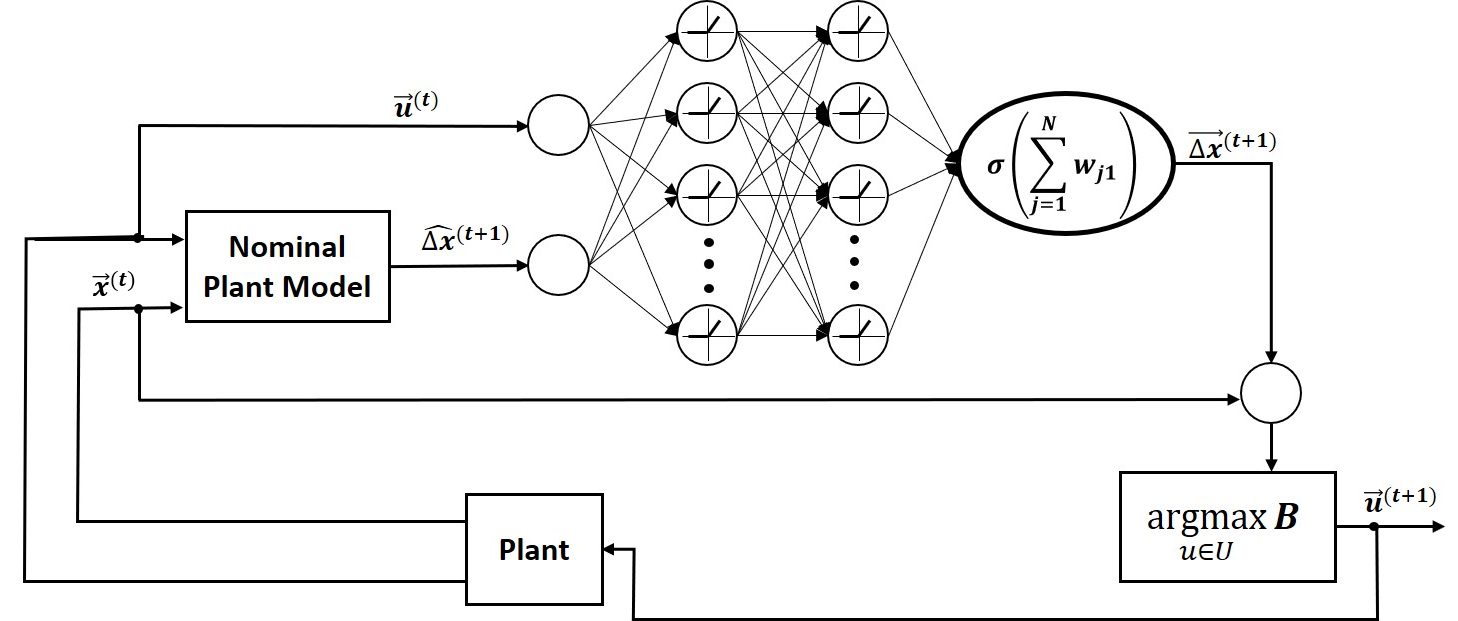}
    \caption{This figure depicts our training architecture where $B$ refers to the criteria that is being maximized in Eq.~\ref{eq:obj}.}
    \label{fig:neuralNetwork}
\end{figure}

In Fig.~\ref{fig:neuralNetwork}, we provide a graphical depiction of our neural, model-learning subroutine. We note that we show the architecture that we utilize to train the neural network. Instead of learning the plant model from scratch, we choose to learn the change in the plant dynamics (the difference between the dynamics estimated by the nominal network or undamaged plant and the true dynamics of the damaged aircraft) via a neural network.  In practice, we find it requires less training examples to learn the change in model dynamics rather than relearn the model from scratch.  Thus, our goal is to learn the mapping of the nominal estimated plant dynamics to the true damaged dynamics. This approach is in keeping with prior work in approximating dynamical models \cite{Kory,Saver,Latimer1977}.

\subsection{Active Learning: Acquiring Additional Information}
\label{sec:AL}

We employ active learning to determine which action provides the most information about the damage. Researchers \cite{Cai2017,Hastie2017} have proposed various acquisition functions (i.e., heuristics that exist to determine which training data point to choose). We explore such functions for learning a damaged aircraft model and introduce our own, well-suited for maximizing information quickly under computational limitations.

\subsubsection{Baseline Acquisition Functions}

\textbf{Model Change }\cite{Cai2017} --  Model change is a measure of the difference between the current model parameters and the updated model parameters after the addition of a training sample. Model change is a good measure of how much the model will have ``learned" after a new training sample is added. We employ the method proposed by \cite{Cai2017} and shown in Eq.~\ref{eq:modelChange}.  The expected model change is defined as the change in weights, $\frac{\delta L_x(\theta)}{{\delta \theta}}$, given a candidate input and associated label x, weighted by the conditional probability of $x$. $\theta$ are the network weights.\par\nobreak{\parskip0pt \footnotesize \noindent\begin{equation}
u^*=\argmax_{u \in \mathbb{U}} \int_X \norm{\biggl. \frac{\delta L(\theta)}{{\delta \theta}} \biggr. } P(x|u) dx \label{eq:modelChange}
\end{equation}}

\noindent\textbf{Epistemic Uncertainty} \cite{Hastie2017} --  We also consider maximizing epistemic uncertainty. Uncertainty of a model can be estimated via the variance of an ensemble of bootstrapped networks. Z ensembles are created via sampling with replacement of the original training data. The variance is calculated as a function of the difference between the outputs of the bootstrapped models, $f_z$ and the average  of the models, $\bar{x}$  as shown in Eq.~\ref{eq:Epistemic}. We choose the candidate which maximizes the variance between the bootstrapped models.\par\nobreak{\parskip0pt \footnotesize \noindent
\begin{equation}
u^*= \argmax_{u \in \mathbb{U}} \frac{1}{Z}\sum_{z=1}^Z(\bar{x}-f_z(u))^2 \label{eq:Epistemic}
\end{equation}}

\subsubsection{Our Acquisition Function}
\noindent\textbf{Maximizing Diversity}-- We propose a novel acquisition function for active model learning defined in Eq.~\ref{eq:I} of Sec.~\ref{sec:ourDesc}. This function minimizes similarity of the candidate data and the predicted output versus the previous training inputs and outputs. We want our model to learn the dynamics of the aircraft across the full range of possible states and actions. Furthermore, we want a computationally light function for fast learning. Training inputs and outputs that differ greatly from those already seen, will provide the most information. 

 \section{Experimental Results}
 \label{sec:results}
 
{\centering
\begin{figure*}
\begin{subfigure}{0.33\textwidth}
  \centering
  \includegraphics[height=4cm]{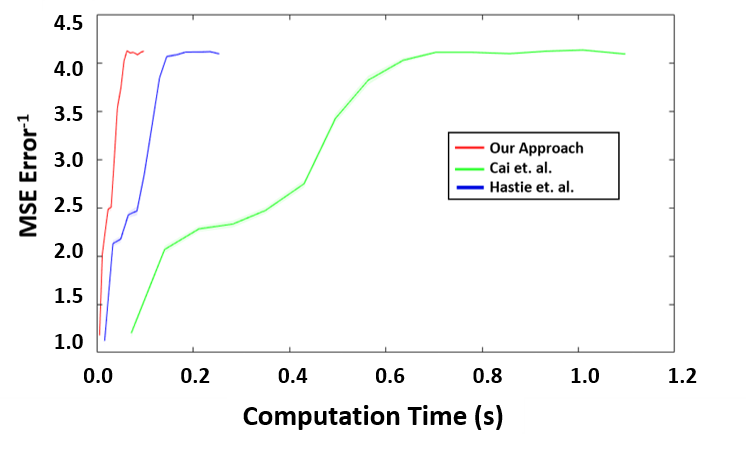}
  \caption{}
  \label{fig:activeLearning}
\end{subfigure}
\begin{subfigure}{0.33\textwidth}
  \centering
  \includegraphics[height=4cm]{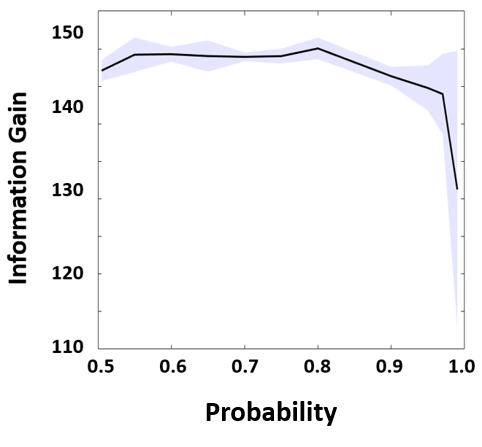}
  \caption{}
  \label{fig:lambdaPlot}
\end{subfigure}%
\begin{subfigure}{0.33\textwidth}
\includegraphics[height=4cm]{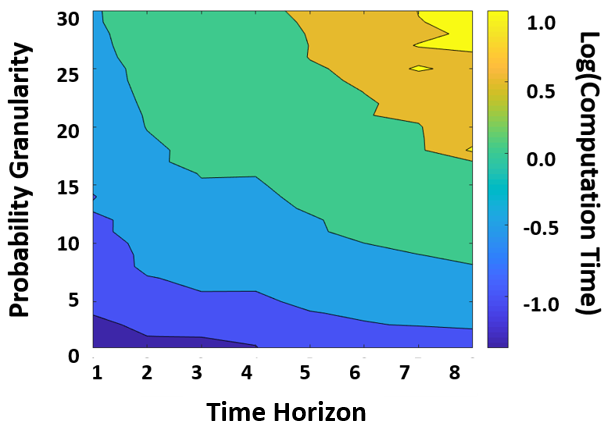}
    \caption{}
        \label{fig:Heat}
\end{subfigure}
\caption{Fig.~\ref{fig:activeLearning} depicts the information gain versus total computational time for increasing number of training samples (N) for each of the three active learning functions. The shaded region depicts the standard error. This analysis was done within the context of our framework. Fig.~\ref{fig:lambdaPlot} depicts the trade-off between probability of safe flight versus acquiring new information. Fig.~\ref{fig:Heat} depicts computation time as a function of the time horizon, $T$, and the number of probability levels, $\delta_{\epsilon}$. Our algorithm can produce safe, information-rich trajectories in $ \frac{1}{10}^{th}$ second for $T=5$ and with $5$ levels of safety.}
\label{fig:test2}
\end{figure*}}

We empirically investigate our bounded rationality, safe framework for MPC. First, we compare the relative merits of the acquisition functions proposed for the active learning (Sec.~\ref{sec:compareAF}). Second, we evaluate the efficacy of our framework for quickly regaining high-functioning control of aircraft under various failure conditions (Sec.~\ref{sec:ultimateEval}). We provide a video supplement demonstrating our simulated damage scenarios. It can be viewed at https://tinyurl.com/y69stkx9. The code for the simulation can be viewed at https://tinyurl.com/y4exh7b4.

\subsection{Comparison of Acquisition Functions}
\label{sec:compareAF}
Fig.~\ref{fig:activeLearning} depicts the improvement in information gain and computation time of our acquisition function compared to the baselines for increasing number of samples, N. We choose to use similarity as our acquisition function since the metric calculations are faster, its linearity is well-suited for our optimization formulation and  it performs at par with other state of the art acquisition function within our framework in terms of information gain.

\subsection{Safe Recovery from Failure}
\label{sec:ultimateEval}

We test our algorithm in a simulated environment on three damage scenarios of a Boeing 747 aircraft: 1) 33\% loss of the left wing, 2) complete loss of vertical stabilizer and rudder, and 3) loss of aileron control. For these damage scenarios, we draw upon prior work that developed theoretical damage models \cite{Zhang2017,Ouellette2010}. Specifically, \cite{Watkiss1994} proposes a 3D aerodynamic state space perturbation model of 33\% loss of the left wing.  \cite{Ogunwa2016} provides a model for complete loss of the vertical stabilizer. The full equations of motion can be found in the cited work. We utilize these perturbed equations of motion to simulate the dynamics of the damaged aircraft. The states of the aircraft are forward velocity ($u$), vertical velocity ($w$), pitch rate ($q$), pitch angle ($\theta$), sideslip angle ($\beta$), roll rate ($p$), yaw rate ($r$), roll angle ($\psi$), yaw angle ($\phi$), lateral coordinate positions ($X$, $Y$), and altitude ($Z$). The control inputs are elevator ($\Delta_e$), thrust ($\Delta_t$), aileron ($\Delta_a$), and rudder ($\Delta_r$).

\begin{figure}[h]
\centering
  \includegraphics[width = 0.65\columnwidth, height = 2.15cm]{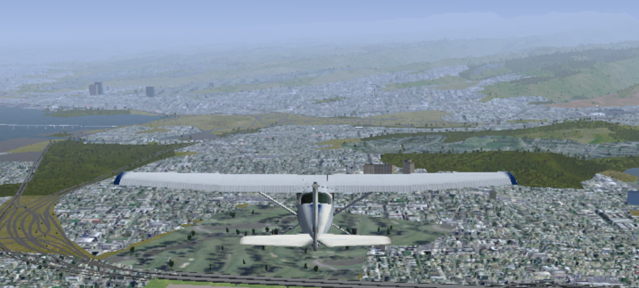}
  \caption{The Flightgear virtual environment.}
  \label{fig:plane}
\end{figure}

We implement our simulation in Simulink using the open source flight simulator, FlightGear (Fig.~\ref{fig:plane}). We simulate sensors (accelerometers, gyroscopes, and magnetometer) with conservative levels of noise and a sampling rate of 20 Hz.  The goal is to learn the dynamics while staying within the safe region to avoid unrecoverable configurations of the aircraft. Given plant inputs $\vec{u}$, states $\vec{x}_k$, and resultant states $\vec{x}_{k+1}$, we learn the model that maps these inputs to outputs. There are eleven states and four controls to the airplane plant, so our neural network must learn a mapping from fifteen inputs to eleven outputs. We choose a simple neural network structure to keep computational time at a minimum.  We found that a single layer perceptron with linear activation function proved adequate to represent the dynamics of the plant. Once the model is learned, we want the aircraft to maintain stable flight, meaning it retains its altitude and zero degrees roll angle. These parameters are chosen to avoid stall or spin scenarios. In our experiment, the desired flight trajectory after damage had occurred was 50 m/s, 0 degrees roll and pitch, and 300 m altitude.  The radius of safety is +/- 10 degrees in the roll and pitch and +/- 30 meters in z away from the reference trajectory \cite{Carbaugh}.  We ran Monte Carlo simulations over starting configurations, i.e., various forward velocities and injected random noise into the environment.
\begin{figure*}
\begin{subfigure}{0.33\textwidth}
  \centering
  \includegraphics[width=\columnwidth]{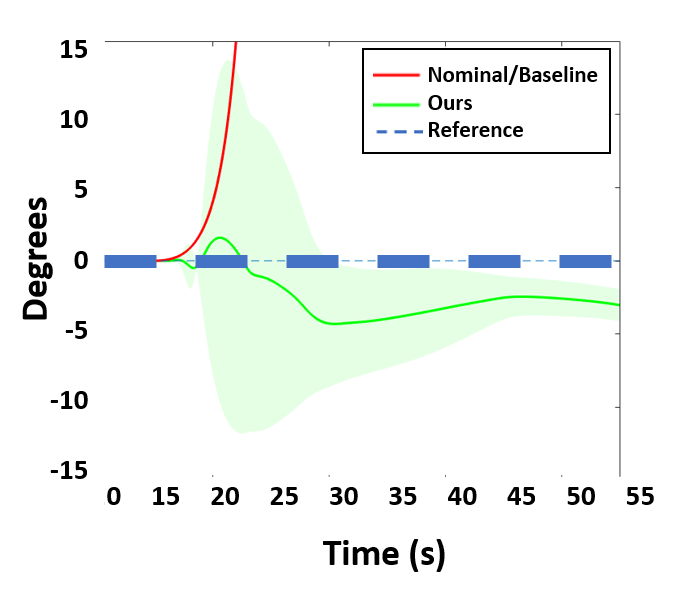}
  \caption{}
  \label{fig:trajectoriesWingRoll}
\end{subfigure}%
\begin{subfigure}{0.33\textwidth}
  \centering
  \includegraphics[width=\columnwidth]{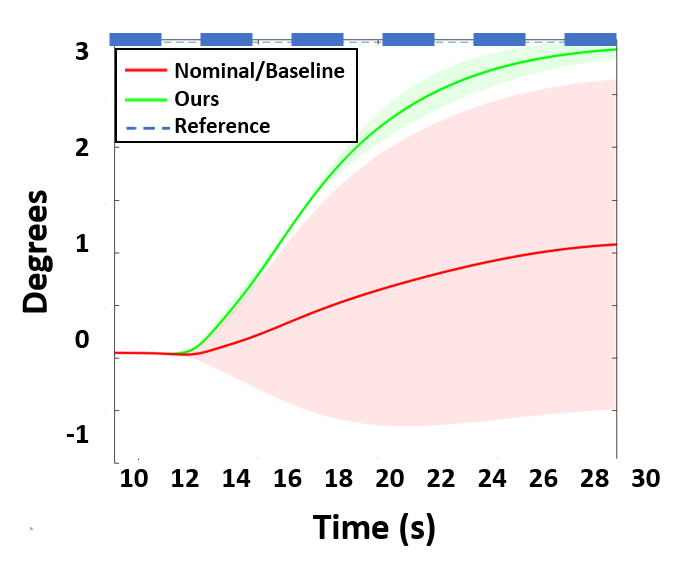}
  \caption{}
  \label{fig:trajectoriesFault2}
\end{subfigure}
\begin{subfigure}{0.33\textwidth}
  \centering
  \includegraphics[width=\columnwidth]{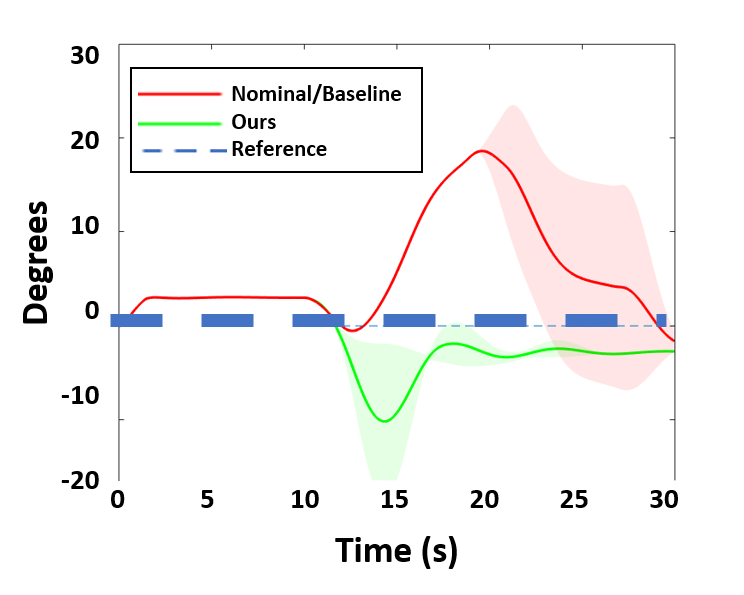}
  \caption{}
  \label{fig:trajectoriesAileron}
\end{subfigure}
\caption{Fig.~\ref{fig:trajectoriesWingRoll} depicts the deviation of roll angle from reference trajectory of zero degrees when wing has been damaged under Failure Condition 1. Fig.~\ref{fig:trajectoriesFault2} depicts the yaw angle of plane when tracking a 3 degree reference trajectory with complete loss of vertical stabilizer and rudder under Failure Condition 2. Fig.~\ref{fig:trajectoriesWingRoll} depicts the deviation of roll angle from reference under Failure Condition 3. The shaded regions depict the variance in the trajectories over the Monte Carlo simulations.} Results are shown for no active learning (baseline) versus our approach.
\label{fig:test}
\end{figure*}

We benchmark our algorithm against a standard MPC. While we would like to benchmark against other active learning frameworks such as those presented in \cite{Bongard2006} and \cite{Cully2015}, their reported time for learning the model are far too slow for regaining control in the few seconds a robot might have before entering a terminal state (e.g., an autonomous car crashing or an airplane entering an unrecoverable spin). The architectures developed in these works require up to minutes to update the damage model, yet an aircraft model must be learned in a few seconds for there to be hope of recovery, especially if the damage is severe. Furthermore, these architectures do not include a notion of safety which is critical when dealing with an unstable system such as an aircraft nor do they account for as wide a variety of damage scenarios as our architecture. Therefore, we compare our algorithm's performance against a nominal MPC.


\textbf{Condition 1) Wing Damage} --  Fig.~\ref{fig:trajectoriesWingRoll} shows that the plane stayed within the desired range under our Monte Carlo simulations and began tracking the reference trajectory once the damage model had been learned. We are able to detect when damage had occurred to the wing within 0.1 seconds and learn an updated plant model in  an additional 5 seconds.  The plane is then controlled using the MPC scheme with the updated plant model once the confidence in the model reaches the specified threshold. It took approximately ten seconds for the aircraft to regain stable flight. However, the roll angle remains at about -3 degrees to compensate for loss of the left wing. Fig.~\ref{fig:trajectoriesWingRoll} shows the improvement attained in performance with active learning.  Only with an active learning strategy, the new plant model was able to be learned in time and the aircraft control was able to be recovered.  When an active learning strategy was not used, the model could not be learned before loss of control.



\label{sec:trajectories2}
\textbf{Condition 2) Loss of Vertical Stabilizer and Rudder} -
In the case of complete loss of vertical stabilizer and rudder, we want to determine if we can safely learn the new dynamics of the plane and track a yaw angle of three degrees.  The dynamics involved with loss of the vertical stabilizer and rudder proved to be an easier model to learn compared to wing loss. This is likely due to the fact that loss of the vertical stabilizer only effects the lateral dynamics of the plane. The network did not have to learn any change between the nominal longitudinal dynamics and the damaged longitudinal dynamics.  Fig.~\ref{fig:trajectoriesFault2} shows that when active learning is utilized, the plane is able to quickly track a 3 degree yaw angle despite loss of rudder. Without the active learning strategy, the plane takes considerably longer to track the reference.




\label{sec:trajectories2}
\textbf{Condition 3) Loss of Aileron Control} -- The last damage scenario presented is complete loss of control of the aileron.  This effects the ability to independently control the roll of the aircraft. The results in Fig.~\ref{fig:trajectoriesAileron} show the speed with which the damaged aircraft reaches the reference trajectory when active learning is used versus when the nominal MPC policy is used.  Because the dynamics of the aircraft are learned more efficiently when utilizing active learning, the aircraft is able to regain stable flight and converge to the reference trajectory considerably faster. 

\subsection{Bounded Rationality Trade-off}
Our bounded rationality framework trades-off acquiring information to improve model accuracy while also maximizing the likelihood of safely accomplishing the task (e.g., safe flight). This trade-off is weighed by a hyper-parameter, $\lambda$. To investigate the sensitivity of our model to this trade-off, we performed a Monte Carlo analysis sweeping $\lambda$ (Fig.~\ref{fig:lambdaPlot}). The resulting curve shows our optimization algorithm is able to achieve a high-probability of safe maneuvering while actively seeking out information to  adapt our dynamics model. We achieve an average $95\%$ chance of safe flight losing $<10\%$ information when discounting safety (i.e., $50\%$ safety).

\subsection{Computation Time}
We investigate the speed of our bounded rationality algorithm. Specifically, we conduct a Monte Carlo sweep of scenarios for various time horizons in planning for $T \in {0,1,\ldots,7}$ and the number of discrete safety settings, $\delta_{\epsilon}$. We find that our algorithm can produce safe, information-rich trajectories in 0.1 seconds for $T=5$ and with $5$ levels of safety as depicted in Fig \ref{fig:Heat}.


\subsection{Sensitivity Analysis}
\label{sec:sensitivity}
We conduct a sensitivity analysis based on \cite{Vahid} conducting Monte Carlo simulations varying model parameters, $\lambda$ and $N$ and calculate the sensitivity of each parameter in relation to each perturbed parameter. The nominal parameters of our model are $\lambda=0.1$ and $N=3$. We vary $\lambda \in \{.001, .01, .1, 1, 10\}$ and $N \in \{0,1,2,3,4,5\}$. In Eq.~\ref{eq:sens1}-\ref{eq:sens2}, $D_j^{\theta}$ is the percent change in the estimation error between the states, $\bar{S_j^{\theta_o}}$, produced by the nominal parameters, $\theta_o$ and error states, $\bar{S_j^{\theta}}$, produced by the perturbed parameters, $\theta$. $\eta$ is the number of Monte Carlo simulations. \par
\begin{figure}[h]
\centering
  \includegraphics[width=0.9\columnwidth]{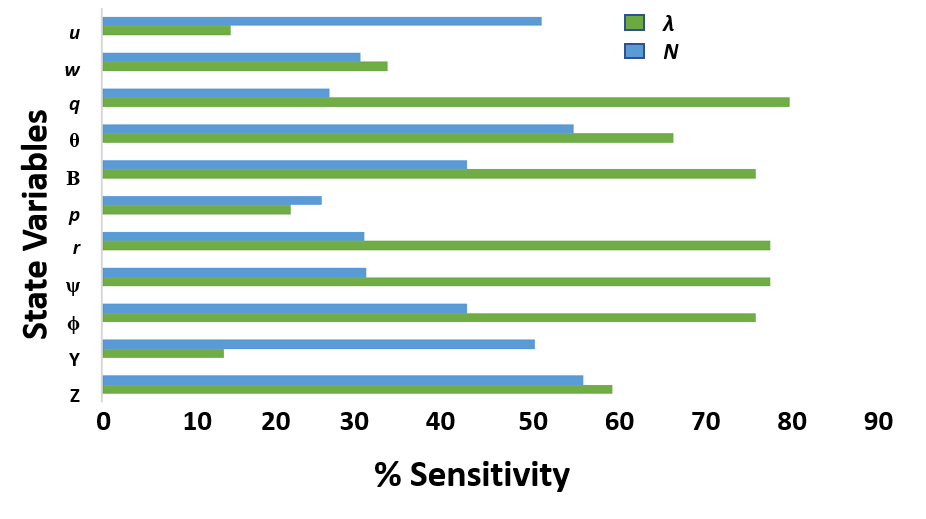}
  \caption{Sensitivity of system to parameter perturbations.}
  \label{fig:sens}
\end{figure}

\nobreak{\parskip0pt \footnotesize \noindent \begin{align}
D_j^{\theta}&=\left(\frac{(max(\bar{S_j^{\theta}})-min(\bar{S_j^{\theta}})}{2}-\bar{S_j^{\theta_0}}\right) \frac{1}{\bar{S_j^{\theta_0}}} \times 100\%
\label{eq:sens1}\\
\bar{S_j^{\theta}}&=\sqrt{\frac{1}{\eta}\sum_{i=1}^N S_{e_i}^{\theta}}
\label{eq:sens2}
\end{align}}\indent Our analysis, shown in Fig \ref{fig:sens}, demonstrates that our control system is robust to deviations in $\theta_o$. Specifically, when holding $N$ at the nominal level, we can vary $\lambda$ a full two orders of magnitude from the nominal amount and only experience between $10\%$-$80\%$ change in our state. Likewise, we can perturb $N$ by a factor $ > 2$ times the nominal level while holding $\lambda$ constant and experience less than a $60\$$ change in our state. These results show that our approach is robust to significant changes in hyper-parameter settings.

\subsection{Summary of Results}
We demonstrate our system's capabilities to recover from a wide variety of damage scenarios and learn the damaged dynamics in sub-second time.  We show that our system learns the dynamics safely, i.e. we are able to guarantee that the aircraft returns to the envelope of safety with a high probability, thus preventing it from reaching an unstable configuration.  Additionally, our active learning framework shows improvement over benchmark metrics in the literature while simultaneously providing a reduced computation time. The combination of our novel active learning framework along with our chance constrained optimization formulation outperforms state-of-the-art model active learning approaches. Our active learning approach is between 19.38\% and 56\% faster than \cite{Hastie2017} and 60.39\% to 78\% faster than \cite{Cai2017} for $N=1$ to $N=15$. Our approach achieves an information gain between .14\% and 7.1\%  greater than \cite{Hastie2017} and between $0.4\%$ and $8.8\%$ greater than \cite{Cai2017} for $N=1$ to $N=15$ as shown in Fig.~\ref{fig:activeLearning}. 

\section{Limitations}
A limitation of our work is that it has only been empirically investigated in the context of fixed-wing aerial vehicles (e.g., UAVs). However, our formulation is designed to be general enough to afford application to systems whose dynamics can be sufficiently approximated by a ReLU neural network encoded within a mathematical program.

Furthermore, our algorithm falls within the vein of mathematical programming-based approaches rather than classical adaptive control schema. As such, we are unable to readily prove that the system is Lyapunov stable. Nonetheless, we show that our system is robust to hyperparameters (Fig.~\ref{fig:sens}) while outperforming state-of-the-art active learning methods (Fig.~\ref{fig:activeLearning}), and we demonstrate that our system is able to safely learn to control a UAV (Fig.~\ref{fig:trajectoriesWingRoll}-\ref{fig:trajectoriesAileron}).”

\section{Future Work: Physical Demonstration}
We provide a brief description of how our system can be verified on a physical fixed wing aircraft. To deploy our system on a physical aircraft, we would utilize a standard bind-n-fly fixed wing aircraft with an on board micro-controller and telemetry. Many drones require low size, weight, power and cost (SWAP) which our system provides. Our system is light-weight enough to be run on a micro-controller such as a Raspberry Pi or Arduino considering our method requires less than 1 MB of memory and utilizes less than 35\% CPU on an average PC.  Damage scenarios can be created by attaching a part of the plane (i.e., part of wing) with solenoids. The power to the solenoids can be cut, thus pushing apart the pre-broken section of the wing. As demonstrated in our simulation, the airplane would utilize the bounded rationality framework, learn the new damage model, and continue to fly. We could monitor the flight and cause damage via telemetry from a ground station. We could test our framework on a variety of damage conditions in this manner.”

\section{Conclusion}
\label{sec:conclusion}
We create a safe, active-learning framework for learning to control a damaged robot. We demonstrate our algorithm's efficacy in simulation for multiple damage scenarios and show our algorithm's ability to maintain safe flight of a UAV. Our novel acquisition function was able to achieve a speed-up of at least $19.38\%$ over prior work, and our algorithm was able to produce safe trajectories in 0.1 seconds. 
As the prevalence of robots and UAVs grows, it is imperative that these systems be equipped with adaptive controllers that can compensate for failures in real time. Our control scheme offers a novel, potential solution to address this challenge.








\bibliographystyle{IEEEtran}
\bibliography{root.bib}

\end{document}